\documentclass[10pt,twocolumn,letterpaper]{article}

\usepackage[pagenumbers]{cvpr}

\usepackage{cuted} 

\usepackage{currfile} 

\usepackage{caption} 

\usepackage{multirow}
\usepackage{array}

\usepackage[T1]{fontenc}
\usepackage{placeins}
\newcommand{\missing}{\textcolor{red}{[?]}}

\definecolor{cvprblue}{rgb}{0.21,0.49,0.74}
\usepackage[pagebackref,breaklinks,colorlinks,allcolors=cvprblue]{hyperref}

\title{GARD-GS: Geometry-Guided Anchor-Regularized Distillation\\from Point Clouds to Gaussian Splatting}

\author{
Puyuan Zhang$^{1}$ \hspace{2em} Jianming Huang$^{2}$ \hspace{2em} Wenkai Ye$^{2}$ \hspace{2em} Wei Dong$^{1}$ \\
[0.65em]
\textnormal{$^{1}$Shanghai Jiao Tong University, Shanghai, China \hspace{2em} $^{2}$Jishu Technology Co., Ltd., China} \\
[0.75em]
\url{https://patrick1159.github.io/gardGS-page/}\\
[0.75em]
}

\begin{document}
\maketitle
\begin{strip}
\vspace{-46pt}
\centering
\includegraphics[width=\textwidth]{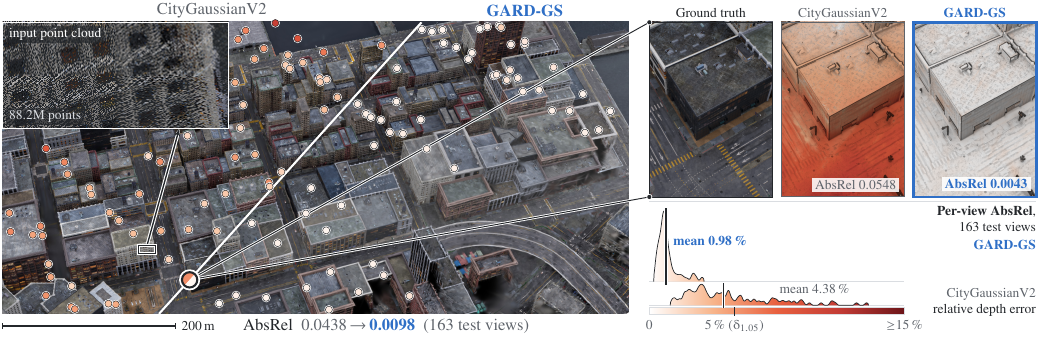}
\captionsetup{hypcap=false}
\captionof{figure}{GARD-GS turns a city-scale point cloud (inset, 88.2M points) into a 3DGS model whose depth follows the measured surface. Left: MatrixCity Aerial Block\_A rendered by CityGaussianV2 (3.83M Gaussians) and by GARD-GS with captured images (0.66M), each test camera colored by its depth AbsRel. Right: at one test view, the relative depth error laid over the ground-truth surface shading; GARD-GS has the lower AbsRel on all 163 test views.}
\captionsetup{hypcap=true}
\label{fig:problem}
\end{strip}

\begin{abstract}
Laser scanning captures city-scale scenes as dense point clouds with centimeter-level accuracy, yet the visual quality of these point clouds rarely matches it.
3D Gaussian Splatting (3DGS) renders in real time with high visual quality but fails to recover accurate geometry, learning it from images through a photometric loss alone.
Existing methods use downsampled point clouds only to initialize or supervise this learning.
This leaves much of the full density and accuracy unused.
In this paper, we present GARD-GS, a method that converts a dense point cloud into a 3DGS model without training or per-scene optimization: each point receives a Gaussian whose covariance comes from the local neighborhood, so the model keeps the measured geometry.
Since a single city block already yields $6.6\times10^7$ Gaussians, GARD-GS simplifies first and optimizes once: merging over many rounds, without any training, reduces the number of Gaussians to 1\%.
A single distillation then recovers the appearance from captured images or, without them, from renderings of the converted 3DGS model, taken at the camera poses of the data or, without poses, at views generated from the point cloud.
The simplified model keeps this geometry: without any captured image, it has less than half the depth error of CityGaussianV2 on both benchmarks and, with captured images, a $4.5\times$ lower depth error on MatrixCity with $5.8\times$ fewer Gaussians.
\end{abstract}

\section{Introduction}
\label{sec:intro}

Dense point clouds are a standard product of 3D capture across industries.
Mobile laser scanning maps urban roadways for surveying and autonomous driving~\cite{tan_toronto3d_2020,roynard_paris_lille_2018,liao_kitti360_2022}, and point clouds also form the geometric basis of digital twins and cultural-heritage records.
Laser scanning reaches centimeter-level or better accuracy~\cite{tan_toronto3d_2020,knapitsch_tanks_2017}, yet the visual quality of point clouds rarely matches it.
The most direct way to view a point cloud, drawing each point as a splat, has long been studied, but the size and shape of each splat remain hard to set, so close-up views show holes or blurred detail~\cite{pfister_surfels_2000,hu_p2enet_2024,liao_point2gaussian_2026}.
Converting a point cloud into a mesh is the usual alternative and suits editing and animation~\cite{guedon_sugar_2024}, but such meshes typically have to be simplified for rendering, and geometric accuracy and texture fidelity are hard to preserve through this reduction~\cite{zwicker_surface_2001}.
A different line of work, radiance fields, learns scenes from multi-view images rather than from point clouds: NeRF~\cite{mildenhall_nerf_2020} represents a scene as a continuous field, and 3D Gaussian Splatting (3DGS)~\cite{kerbl_3d_2023} later replaced this field with explicit anisotropic Gaussians whose positions, scales and rotations adapt to the scene, rendering them in real time with high visual quality.
These Gaussians are explicit like points and meshes, and 3DGS has in turn become a hub for conversion to and from both~\cite{guedon_sugar_2024,zhu_mesh2gs_2026,xiong_nanogs_2026,bendouro_transcoding_2026,hu_p2enet_2024,liao_point2gaussian_2026}.

3DGS learns the geometry of its Gaussians from multi-view images, jointly with their appearance, by optimizing them against a photometric loss~\cite{kerbl_3d_2023}.
This loss alone hardly ensures geometric accuracy~\cite{chen_pgsr_2024}, and the optimization tends to fit every training view while neglecting the underlying scene geometry~\cite{lu_scaffold-gs_2023}.
The Gaussians can therefore reproduce the images well even where their depth is inaccurate~\cite{wu_blockgaussian_2025,fan_trimgs_2024} (\cref{fig:problem}).
Such depth errors are rarely measured, and at city scale only recently~\cite{liu_citygsv2_2025}.
A dense point cloud, by contrast, measures the scene surfaces directly.

Yet 3DGS pipelines mostly use point clouds to help produce a model rather than to become one: point clouds initialize Gaussians that are then optimized against captured images~\cite{guo_pocketgs_2026,hwang_vegs_2024,ha_gsicp_2024,su_gausslam_2025,weng_structuredligs_2026,zhou_drivinggaussian_2024,jiang_li-gs_2024,hess_splatad_2025}, are rendered into views that supervise the same optimization when no images exist~\cite{bendouro_transcoding_2026,liao_point2gaussian_2026}, or feed networks that predict Gaussians~\cite{hu_p2enet_2024,zhou_gaussianpainter_2024}.
Converting a point cloud directly into a 3DGS model has rarely been attempted, although point-based rendering has long derived each splat from the neighborhood of its point~\cite{zwicker_surface_2001,hu_p2enet_2024} and point cloud registration likewise fits a Gaussian to the neighborhood of each point~\cite{segal_gicp_2009}.

Such a conversion, however, yields one Gaussian per point, about $6.6\times10^7$ for a MatrixCity aerial block and more than ten times as many as a 3DGS model trained on its images (\cref{tab:main}).
Image-free simplification~\cite{xiong_nanogs_2026,li_cvtgs_2026} lowers such counts and applies to converted 3DGS models, yet typically blurs fine texture at high simplification ratios.
Image-based methods usually recover this detail from the captured images, directly or through a teacher fitted to them~\cite{fan_lightgaussian_nodate,xiang_distilled-3dgsdistilled_2025}, but captured images of a laser-scanned area may be missing, sparse, or confined to the capture trajectory.
Without such images, the converted 3DGS model serves as the teacher, and its renderings become the only supervision~\cite{bendouro_transcoding_2026}.
A large converted 3DGS model therefore needs a simplification that keeps its geometry and a recovery stage that works from the point cloud alone.

We thus address a different task: turning a dense measured point cloud, with or without captured images, into a lightweight 3DGS model that keeps the measured geometry.
In this paper, we present GARD-GS, a method that converts a dense point cloud into a 3DGS model, simplifies this model, and recovers its appearance (\cref{fig:overview}).
The conversion places a Gaussian at every point and computes its covariance from the local neighborhood, so the converted 3DGS model takes its geometry from the measured points alone, with $17\times$ lower depth error than a 3DGS model trained on the same aerial images.
Simplification then merges neighboring Gaussians~\cite{xiong_nanogs_2026} over many rounds without any training and, unlike iterative pruning~\cite{hanson_pup_2025}, defers all optimization to a single distillation at the end.
This distillation restores the appearance from captured images when they exist and otherwise from renderings of the teacher, taken at the camera poses provided with the data or, where no poses exist, at views generated from the point cloud and selected to cover its surface~\cite{li_magician_2026}.

We evaluate GARD-GS on MatrixCity~\cite{li_matrixcity_2023} and GauU-Scene~\cite{xiong_gauuscene_2024}.
Our contributions are:
\begin{itemize}
  \item \textbf{Conversion.} A direct conversion of large-scale point clouds into a 3DGS model, which places a Gaussian at every point and computes its covariance from the local neighborhood without learning or per-scene optimization, and yields more accurate geometry than 3DGS models trained on images.
  \item \textbf{Simplify first, optimize once.} A simplification scheme for converted models that reduces the number of Gaussians by about $100\times$ without any training and defers all optimization to a single distillation at the end.
Fitting between merging rounds, as in prune-and-refine schemes, brings no gain in geometry at about four times the optimization steps.
  \item \textbf{Geometry-preserving distillation.} A distillation that restores the appearance of the simplified model from captured images or, when none exist, from renderings of the converted model.
It starts from Gaussians on the measured surfaces and adds none, so it keeps the input geometry rather than re-estimating it, and yields lower depth error than every method trained on images.
\end{itemize}

\section{Related Work}
\label{sec:related}

\subsection{Point Clouds in Gaussian Splatting}
\label{sec:rw_points}

\begin{figure*}[t]
\centering
\includegraphics[width=\textwidth]{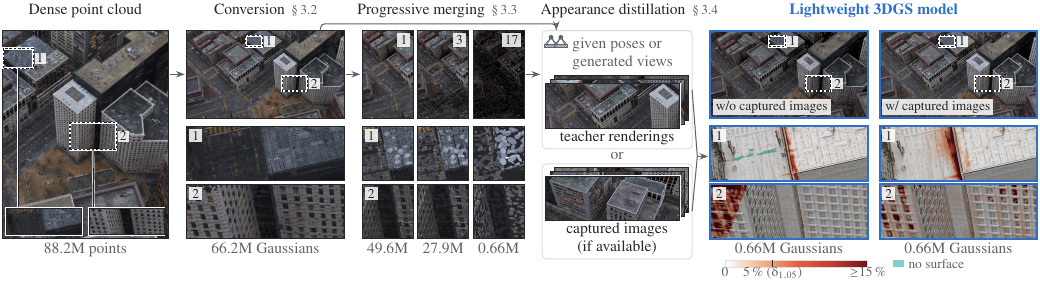}
\caption{GARD-GS turns a dense point cloud into a lightweight 3DGS model through training-free conversion, progressive merging and appearance distillation; captured images are optional. The top row shows the full view, and boxes 1 and 2 mark close-ups from test views, where the outputs show the depth error against ground truth, over the ground-truth surface shading (capped at 15\%, 5\% marked). At 0.66M Gaussians, the final model has a lower AbsRel than the 66.2M converted model (0.024): 0.020 without and 0.010 with captured images.}
\label{fig:overview}
\end{figure*}

Since 3D Gaussian Splatting (3DGS)~\cite{kerbl_3d_2023} initialized its Gaussians from the sparse point cloud of structure-from-motion, point clouds have mostly served as a starting point for its photometric optimization.
LiDAR point clouds later replaced the structure-from-motion point cloud in driving, urban and aerial scenes, usually together with additional depth or LiDAR losses~\cite{jiang_li-gs_2024,zhou_drivinggaussian_2024,hess_splatad_2025,hwang_vegs_2024,lim_lidar3dgs_2024,shen_ligsm_2025,yao_arsgaussian_2026,weng_structuredligs_2026,yan_gsslam_2024,su_gausslam_2025}.
Aimed at novel view synthesis, these pipelines typically downsample the LiDAR point clouds~\cite{zhou_drivinggaussian_2024,lim_lidar3dgs_2024,weng_structuredligs_2026}, since overly dense point clouds strain GPU memory~\cite{yao_arsgaussian_2026} and can even interfere with the optimization~\cite{zhou_drivinggaussian_2024}.
Without captured images, renderings of the point cloud supervise the same optimization~\cite{bendouro_transcoding_2026,liao_point2gaussian_2026}, or feed-forward networks predict the Gaussians~\cite{hu_p2enet_2024,zhou_gaussianpainter_2024}.
Point2Gaussian~\cite{liao_point2gaussian_2026} applies such supervision to a laser-scanned point cloud of $3\times10^8$ points, yet it still starts from a downsampled point cloud.
In each of these three roles, as initialization, supervision or network input, the point cloud helps produce the Gaussians, while photometric optimization or a trained network determines their geometry.

Computing a primitive's shape from the neighborhood of its point, rather than by optimization, predates 3DGS: point-based rendering has long sized each splat from nearby points~\cite{pfister_surfels_2000,zwicker_surface_2001,preiner_autosplats_2012}.
Generalized-ICP~\cite{segal_gicp_2009} registers point clouds by treating each point as a Gaussian flattened along its local surface normal, and Voxelized GICP~\cite{koide_vgicp_2021} aggregates these Gaussians within voxels.
Such Gaussians have since been rescaled into a 3DGS map shared by registration and rendering~\cite{ha_gsicp_2024}, while the normals and spacings of nearby points initialize anisotropic Gaussians on phone captures~\cite{guo_pocketgs_2026}.
Building on these works, we convert a dense point cloud into a 3DGS model by computing the scale and rotation of a Gaussian at every point from the covariance of its local neighborhood.
Unlike the 3DGS methods~\cite{ha_gsicp_2024,guo_pocketgs_2026}, which go on to optimize their Gaussians, this conversion keeps every point position of a city-scale point cloud and requires no optimization.

\subsection{Simplification of 3DGS Models}
\label{sec:rw_compact}

\paragraph{Simplification.}
Most compaction methods for 3DGS~\cite{bagdasarian_3dgszip_2025} work while training on captured images~\cite{youn_successgs_2025}: they remove or mask Gaussians that contribute little to the training views~\cite{fang_minisplatting_2024,hanson_speedysplat_2025,ren_fastgs_2026,liu_flexgs_2025,wang_prune_2026,deng_compactgsslam_2024}, quantize or entropy-code their attributes~\cite{girish_eagles_2024,lee_compact3dgs_2024,navaneet_compgs_2024,niedermayr_compressed3dgs_2024,chen_hac_2024}, or organize them around anchors and levels of detail~\cite{lu_scaffold-gs_2023,liu_citygs_2024,shen_lodgs_2025}, also at city scale~\cite{ren_octreegs_2024,gao_proxygs_2025}.
Simplifying a trained model afterward separates compaction from reconstruction, and successive methods rely less and less on captured images.
LightGaussian~\cite{fan_lightgaussian_nodate} and PUP 3D-GS~\cite{hanson_pup_2025} score Gaussians on the training views and then fine-tune on those images.
GHAP~\cite{wang_ghap_2025} reduces the Gaussians without images before fine-tuning their color and opacity on the training images, and recent methods simplify without images at all~\cite{li_cvtgs_2026,tran_pointsplat_2026,mousa_coreset_2026}.
Among them, NanoGS~\cite{xiong_nanogs_2026} merges neighboring Gaussians without training, motivated in part by Gaussians converted from meshes or point clouds.
These methods, however, compress 3DGS models trained on captured images and judge them by image metrics on the same capture.
For a 3DGS model converted from a point cloud, which holds one Gaussian for every measured point, simplification is an indispensable step after conversion rather than an optional compression, given its tens of millions of Gaussians.
This simplification must also keep the measured geometry, whether or not captured images exist.

\paragraph{Appearance recovery and view selection.}
After simplification, appearance is typically recovered by distillation from a teacher fitted to captured images and rendered at or near the training views~\cite{fan_lightgaussian_nodate,xiang_distilled-3dgsdistilled_2025}, or at pseudo-views warped or generated from them~\cite{wu_blockgaussian_2025,wu_difix3d_2025,hwang_vegs_2024}.
Without captured images, the input itself serves as the teacher and is rendered from cameras preset around each object or part~\cite{bendouro_transcoding_2026,liao_point2gaussian_2026,liao_mega_2026}.
These renderings, the only supervision, cover just the surfaces their views observe, so the choice of views decides what can be recovered.
Active reconstruction makes this choice of views for image capture, ranking candidate poses by uncertainty~\cite{pan_activenerf_2022,jiang_fisherrf_2024} or by the poorly observed surface they cover~\cite{li_magician_2026,chen_cover_2026}.
We borrow the coverage criterion to choose the views at which the teacher is rendered.

\subsection{Geometry-Accurate Gaussian Splatting}
\label{sec:rw_geometry}

Geometry-accurate variants of 3DGS start from the observation that a photometric loss alone hardly ensures accurate geometry. 2DGS~\cite{huang_2dgs_2024} flattens Gaussians into disks and regularizes their depth and normals.
Other regularizers pursue the same aim~\cite{chen_pgsr_2024,guedon_sugar_2024,fan_trimgs_2024,hyung_erank_2024,chen_fds_2025}, and CityGaussianV2~\cite{liu_citygsv2_2025} carries it to city scale with monocular depth.
All of these methods recover geometry from captured images and take no dense point cloud as input.
Once such a point cloud exists, the geometry is already measured: it is no longer an unknown to estimate but a given to extract and keep.

\section{Method}
\label{sec:method}

\begin{figure}[t]
\centering
\includegraphics[width=\linewidth]{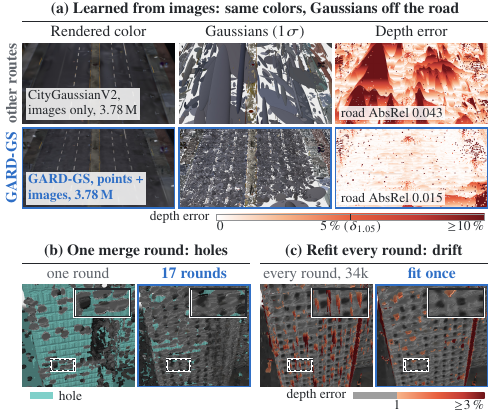}
\caption{Three ways to lose geometry. (a) CityGaussianV2, trained on images, renders the street correctly, yet its Gaussians leave the road; GARD-GS keeps its Gaussians on the road (3.78M each). (b) Before distillation, one merge round leaves holes (teal) that 17 rounds avoid. (c) Fitting after every round leaves bands of drifted depth on the facade; GARD-GS, fitted once, keeps the surface.}
\label{fig:pain}
\end{figure}

Given a dense point cloud of a large scene, our goal is a 3DGS model that preserves the geometric accuracy of the point cloud and renders reliable appearance, whether or not captured images exist.
This accuracy shows in the rendered depth, which the positions, covariances and opacities of the Gaussians determine (\cref{sec:prelim}).
A photometric loss supervises only the rendered colors, and since a large Gaussian far from the camera can render the same pixels as a small one close to it, optimizing the Gaussians against images leaves the depth weakly constrained~\cite{chen_pgsr_2024,fan_trimgs_2024} (\cref{fig:pain}a).
Converting the point cloud instead makes the measured geometry part of the model from the start: each Gaussian takes its position from a point and its covariance from the neighborhood of that point (\cref{sec:conversion}; \cref{fig:overview}).

The converted 3DGS model, however, inherits the sampling density of the point cloud, far beyond what rendering requires.
A second problem is therefore to simplify this model and to restore the appearance that simplification averages away, while keeping the geometric accuracy of the conversion.
Without access to the measured geometry, both Gaussian simplification (which aggregates local Gaussian groups) and color-only fitting tend to pull the rendered depth away from that of the converted 3DGS model, even when applied just once (\cref{sec:exp_ablation}).
We call this deviation of the rendered depth from the converted 3DGS model \textbf{geometric drift} (\cref{fig:pain}b,c).

\subsection{Preliminaries}
\label{sec:prelim}

Following 3DGS~\cite{kerbl_3d_2023}, a 3DGS model $\mathcal{G}=\{G_i\}_{i=1}^{N}$ consists of $N$ Gaussians.
Each Gaussian $G_i$ has a mean $\boldsymbol{\mu}_i$, an opacity $\alpha_i$, a color $\mathbf{c}_i$, and a covariance $\boldsymbol{\Sigma}_i=\mathbf{R}_i\mathbf{S}_i\mathbf{S}_i^{\top}\mathbf{R}_i^{\top}$ with rotation $\mathbf{R}_i$ and scales $\mathbf{S}_i=\mathrm{diag}(\mathbf{s}_i)$ along its three axes.
For a camera $\phi$, the Gaussians that overlap a pixel $\mathbf{x}$ are sorted by depth and blended,
\begin{equation}
C(\mathbf{x})=\sum_{i}\mathbf{c}_i\,w_i(\mathbf{x}),\qquad
D(\mathbf{x})=\frac{\sum_{i}z_i\,w_i(\mathbf{x})}{\sum_{i}w_i(\mathbf{x})},
\end{equation}
where $\tilde{\alpha}_i(\mathbf{x})$ is the opacity $\alpha_i$ times the projection of $G_i$ evaluated at $\mathbf{x}$, and $z_i$ is the depth of $\boldsymbol{\mu}_i$ in the camera.
The weight $w_i(\mathbf{x})=\tilde{\alpha}_i(\mathbf{x})\,T_i(\mathbf{x})$ scales $\tilde{\alpha}_i$ by the transmittance $T_i(\mathbf{x})=\prod_{j<i}(1-\tilde{\alpha}_j(\mathbf{x}))$ in front of $G_i$. $C$ is the rendered color and $D$ the expected depth.
The median depth $D_{\mathrm{med}}(\mathbf{x})=\max\{z_i \mid T_i(\mathbf{x})>0.5\}$ is the largest depth still visible~\cite{huang_2dgs_2024}.
A 3DGS model trained on images minimizes the photometric loss $\mathcal{L}_{\mathrm{rgb}}$, a combination of $\mathcal{L}_1$ and $\mathcal{L}_{\text{D-SSIM}}$, between $C$ and the images.

\subsection{Conversion}
\label{sec:conversion}

The conversion computes one Gaussian for every point of a dense point cloud $\mathcal{P}=\{(\mathbf{p}_i,\mathbf{c}_i)\}_{i=1}^{M}$ of $M$ points with positions $\mathbf{p}_i$ and colors $\mathbf{c}_i$. Isolated points, typically measurement noise, would become floating Gaussians in front of the scene, so a point is kept only if enough other points lie within a small radius.
For a kept point, the covariance $\boldsymbol{\Sigma}^{\mathcal{N}}_i$ of its $k$ nearest neighbors $\mathcal{N}_i$ describes the local surface.
It factors as $\boldsymbol{\Sigma}^{\mathcal{N}}_i=\mathbf{V}_i\,\mathrm{diag}(\nu_{i,1},\nu_{i,2},\nu_{i,3})\,\mathbf{V}_i^{\top}$ with $\nu_{i,1}\ge\nu_{i,2}\ge\nu_{i,3}$ and orthonormal eigenvectors $\mathbf{V}_i$. This factorization has the form $\mathbf{R}_i\mathbf{S}_i\mathbf{S}_i^{\top}\mathbf{R}_i^{\top}$ of a Gaussian covariance: the eigenvectors give the orientation, $\sqrt{\nu_{i,3}}$ the measured thickness along the normal, and $\nu_{i,1},\nu_{i,2}$ the tangential spread of the neighborhood.
The Gaussian therefore takes $\boldsymbol{\mu}_i=\mathbf{p}_i$, $\mathbf{R}_i=\mathbf{V}_i$ and the scales
\begin{equation}
s_{i,j}=\gamma\sqrt{4\pi\,\nu_{i,j}/k}\quad(j=1,2),\qquad s_{i,3}=\sqrt{\nu_{i,3}},
\end{equation}
where the tangential scales are a fraction $\gamma$ of the local spacing $g$, so that neighboring Gaussians overlap slightly; $k$ neighbors spread evenly over a flat disc give $\nu_{i,j}=kg^2/(4\pi)$. With the color of its point and a constant opacity, each Gaussian is fully determined, and the conversion to $\mathcal{G}_0$ involves no learning or per-scene optimization.

\subsection{Progressive Merging}
\label{sec:merging}

We simplify $\mathcal{G}_0$ to a target number $N_S$ of Gaussians in rounds of training-free merging and defer all optimization to the single distillation of \cref{sec:distillation}.
Unlike in pruning methods for 3DGS models trained on images~\cite{hanson_pup_2025}, no fitting happens between rounds: merging takes the Gaussians as the previous round leaves them, so any drift that fitting caused would be carried by later rounds into ever larger Gaussians.

Pruning and merging complement each other.
Pruning leaves the remaining Gaussians untouched but takes the geometry of the pruned ones with it, whereas merging replaces a group of neighboring Gaussians with one Gaussian computed from its members and so keeps the region that the group covered.
Our operator therefore merges as its main operation and prunes only the Gaussians that exceed the budget of a round after merging.

In each round, the Gaussians of one voxel with similar colors form a group $\mathcal{Q}$, which is replaced by one Gaussian whose mean and covariance are the opacity-weighted moments of its members, as in training-free Gaussian merging~\cite{xiong_nanogs_2026}.
With the normalized weights $\bar{\alpha}_i=\alpha_i/\sum_{j\in\mathcal{Q}}\alpha_j$, the mean is $\boldsymbol{\mu}=\sum_{i\in\mathcal{Q}}\bar{\alpha}_i\boldsymbol{\mu}_i$ and the covariance is
\begin{equation}
\boldsymbol{\Sigma}=\sum_{i\in\mathcal{Q}}\bar{\alpha}_i\big(\boldsymbol{\Sigma}_i+(\boldsymbol{\mu}_i-\boldsymbol{\mu})(\boldsymbol{\mu}_i-\boldsymbol{\mu})^{\top}\big),
\end{equation}
where the second term of $\boldsymbol{\Sigma}$ adds the spread of the member means, so that the merged Gaussian covers its members' region.
Its color is the opacity-weighted mean of the member colors, and its opacity is set so that it blocks as much light as its members together (see the supplementary material).

A merged Gaussian, however, stays close to its members only when they form one compact patch; for a group spanning two surfaces, or the gap between them, it bridges them with surface that the point cloud never measured.
A group is therefore merged only if its members cover at least a fraction $\kappa$ of the merged area, $\sum_{i\in\mathcal{Q}}A_i\ge\kappa A$, with $A_i$ and $A$ the areas that a member and the merged Gaussian cover in their tangent planes.
A group that fails this test is split along its principal axis and tested again.
If a round holds more than $N_r$ Gaussians after merging, we keep the $N_r$ Gaussians with the largest optical mass $m_i=\tau_iA_i$ and prune the rest, where $\tau_i=-\ln(1-\alpha_i)$ is the optical depth of Gaussian $i$ and $A_i$ its area as in the coverage test above.
A pruned Gaussian thus blocks the least light over its area, and merging preserves this optical mass.

Within one round of our merge, a Gaussian joins at most one merge and only compact groups are merged, so a large reduction in a single round would fall mostly to pruning, which empties entire regions once most Gaussians have to be pruned (\cref{sec:exp_multiround}).
Our operator therefore runs in many rounds with a keep ratio $\rho$ close to one.
Starting from $N_0=|\mathcal{G}_0|$, round $r$ leaves $N_r=\max(\lfloor\rho N_{r-1}\rfloor,N_S)$ Gaussians, until $N_r=N_S$. Gaussians merged in one round are merged again later, so groups grow from round to round while every merge stays compact, and pruning takes at most a fraction $1-\rho$ in any round (\cref{fig:multiround}c).
The last round outputs the simplified 3DGS model $\mathcal{G}_S$.

\begin{table*}[t]
\caption{Quantitative comparison of geometry and rendering quality on MatrixCity Aerial and GauU-Scene (three-scene mean). \#GS: number of Gaussians. FPS: RGB rendering at evaluation resolution. CPU merging on MatrixCity Aerial: 1.6\,h (ours), 1.4\,h (NanoGS). $^{\ddagger}$Reported by CityGaussianV2 on A100 GPUs.}
\label{tab:main}
\centering
\footnotesize
\sbox0{%
\begin{tabular}{@{}clcccccccc@{}}
\toprule
 & Method & F1$\uparrow$ & AbsRel$\downarrow$ & $\delta_{1.05}$$\uparrow$ & PSNR$\uparrow$ & \#GS & Size & FPS & GPU$\cdot$h \\
\midrule
\multirow{8}{*}{\rotatebox[origin=c]{90}{MatrixCity Aerial}} & 3DGS & 0.291 & 0.4168 & 0.281 & 20.75 & 5.44M & 1.28 GB & 181 & 0.83 \\
 & CityGaussianV2 & 0.527 & 0.0438 & 0.706 & \textbf{27.43} & 3.83M & 567 MB & 27.0$^{\ddagger}$ & 24.1$^{\ddagger}$ \\
\cmidrule{2-10}
 & NanoGS & 0.617 & 0.0208 & 0.897 & 17.34 & 0.66M & 37.1 MB & 1111 & 0 \\
 & LightGaussian$^{\dagger}$ & 0.029 & 0.2265 & 0.173 & 18.45 & 0.66M & 45.0 MB & 731 & 0.20 \\
 & PUP 3D-GS$^{\dagger}$ & 0.466 & 0.0539 & 0.646 & 25.02 & 0.66M & 45.0 MB & 893 & 0.59 \\
\cmidrule{2-10}
 & Ours, converted & \underline{0.775} & 0.0244 & 0.937 & 19.56 & 66.2M & 3.71 GB & 61 & 0.01 \\
 & Ours, w/o captured images & \textbf{0.777} & \underline{0.0198} & \underline{0.944} & 21.36 & 0.66M & 37.2 MB & 833 & 0.14 \\
 & Ours, w/ captured images & 0.757 & \textbf{0.0098} & \textbf{0.968} & \underline{25.28} & 0.66M & 37.2 MB & 652 & 0.17 \\
\midrule
\multirow{7}{*}{\rotatebox[origin=c]{90}{GauU-Scene}} & CityGaussianV2 & 0.391 & 0.0637 & 0.641 & 16.34 & 7.13M & 1.06 GB & 37.8$^{\ddagger}$ & 24.1$^{\ddagger}$ \\
\cmidrule{2-10}
 & NanoGS & \underline{0.462} & \underline{0.0265} & 0.857 & 14.16 & 0.74M & 41.3 MB & 240 & 0 \\
 & LightGaussian$^{\dagger}$ & 0.061 & 0.2485 & 0.067 & 15.03 & 0.74M & 50.1 MB & 109 & 0.38 \\
 & PUP 3D-GS$^{\dagger}$ & 0.222 & 0.0648 & 0.579 & \underline{16.77} & 0.74M & 50.1 MB & 122 & 1.25 \\
\cmidrule{2-10}
 & Ours, converted & \textbf{0.540} & \textbf{0.0235} & \textbf{0.875} & 15.27 & 73.7M & 4.13 GB & 21.0 & 0.02 \\
 & Ours, w/o captured images & \underline{0.462} & \underline{0.0265} & \underline{0.862} & 15.91 & 0.74M & 41.3 MB & 145 & 0.26 \\
 & Ours, w/ captured images & 0.383 & 0.0347 & 0.820 & \textbf{19.94} & 0.74M & 41.3 MB & 92 & 0.39 \\
\bottomrule
\end{tabular}}
\typeout{TABWIDTH tab:main \the\wd0\space\the\textwidth}
\ifdim\wd0>\textwidth\resizebox{\textwidth}{!}{\usebox0}\else\usebox0\fi
\end{table*}

\subsection{Appearance Distillation}
\label{sec:distillation}

Merging keeps the Gaussians on the measured surfaces but leaves a coarser model: each merged color is the mean of its members, and the Gaussians pruned in each round leave holes.
Distillation restores the appearance and closes these holes in a single photometric optimization of all parameters of $\mathcal{G}_S$, with the number of Gaussians fixed at $N_S$. Each view $\phi$ is supervised by a target image $I^{*}(\phi)$: with captured images, the captured image; without captured images, the rendering $I_{\mathcal{G}_0}(\phi)$ of the converted 3DGS model, which then acts as a teacher~\cite{fan_lightgaussian_nodate,bendouro_transcoding_2026}.
Distillation minimizes
\begin{equation}
\mathcal{L}=\mathcal{L}_{\mathrm{rgb}}\big(I_{\mathcal{G}_S}(\phi),I^{*}(\phi)\big)+\lambda\,\mathcal{L}_{\mathrm{scale}},
\end{equation}
where $\mathcal{L}_{\mathrm{scale}}$ is a scale regularizer weighted by $\lambda$. Starting from Gaussians that already lie on the measured surfaces and adding none, distillation keeps the geometry inherited from the converted 3DGS model rather than re-estimating it (\cref{sec:exp_comparison}).

\paragraph{Scale regularizer.} Under the photometric loss alone, a few Gaussians grow far beyond any converted Gaussian to cover pixels that many small ones covered, so the model overfits the supervising views while its depth leaves the surface in other views.
The growth is strongest with captured images, since they also show content that the point cloud never measured, such as the sky and distant background.
The regularizer is therefore used only with captured images, where it penalizes the largest scale $s_i^{\max}=\max_j s_{i,j}$ of a Gaussian beyond a bound $\bar{s}$,
\begin{equation}
\mathcal{L}_{\mathrm{scale}}=\operatorname*{mean}_{i:\,s_i^{\max}>\bar{s}}\Big(\log\frac{s_i^{\max}}{\bar{s}}\Big)^{2}.
\end{equation}

\paragraph{Views without camera poses.} With the point cloud only, the teacher is rendered at views generated above the point cloud, looking down on every part of the scene that contains points.
From thousands of such views, a subset is selected greedily to cover the observed surface from varied directions and distances (see the supplementary material).

\section{Experiments}
\label{sec:exp}

\subsection{Experimental Setup}
\label{sec:exp_setup}

\paragraph{Datasets.} We evaluate on the aerial views of MatrixCity~\cite{li_matrixcity_2023} and on three scenes of GauU-Scene~\cite{xiong_gauuscene_2024}\footnote{We use version 2.1 of GauU-Scene, whose LiDAR point clouds and images are registered in a metric geographic frame.} (Residence, Russian Building and Modern Building), the two large-scale benchmarks of CityGaussianV2~\cite{liu_citygsv2_2025}.
Both provide a dense point cloud, captured images registered to it, and geometric ground truth.
On MatrixCity, the official aerial point cloud of Block\_A holds $8.8\times10^7$ points, and GauU-Scene aligns airborne LiDAR point clouds with drone images.

\paragraph{Evaluation protocol.} In our setting, the task is to keep the geometry that the scan measured rather than to re-estimate it, so all methods are scored against the same scan, as CityGaussianV2 does on GauU-Scene.
At every test view, we render the expected depth of each model (the alpha-blended depth $D$ of \cref{sec:prelim}, so any Gaussian off the surface raises its error) and report the mean absolute relative depth error (AbsRel) and $\delta_{1.05}$, the fraction of pixels whose depth lies within 5\% of the ground truth~\cite{eigen_depth_2014,beche_claravid_2025}, computing both ourselves for every model.
Pixels where a model renders no surface count as failures, which penalizes holes.
Meshes are extracted and scored following the protocol of CityGaussianV2, with one set of parameters shared by every model: TSDF fusion of the rendered median depth at all training views, ICP alignment to the ground truth, and F1 against the ground-truth point cloud.
PSNR, without the sky on GauU-Scene, serves as a secondary metric.

\begin{figure*}[t]
\centering
\includegraphics[width=\textwidth]{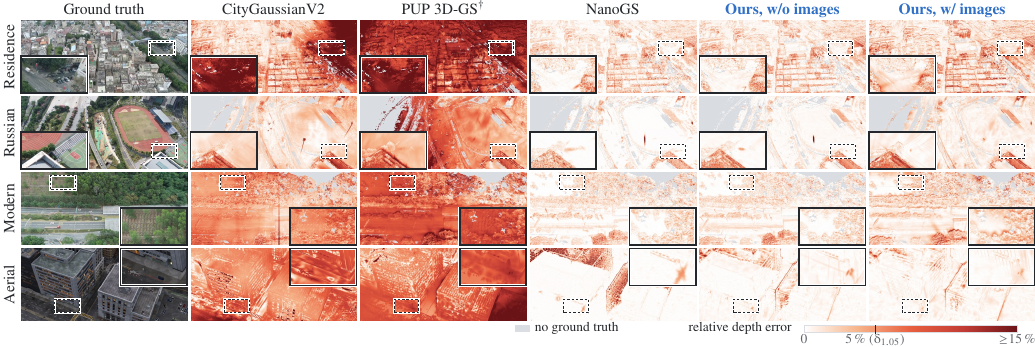}
\caption{Per-pixel relative error of the expected depth on GauU-Scene and MatrixCity Aerial. Color runs from white to red (capped at 15\%, 5\% marked); light gray pixels have no ground truth; dashed boxes are enlarged 2.7 times. CityGaussianV2 and PUP 3D-GS$^{\dagger}$ leave the ground-truth surfaces, whereas GARD-GS, without or with captured images, stays on them.}
\label{fig:qual}
\end{figure*}

\paragraph{Baselines.} Of the methods that learn the geometry from images, we train 3DGS~\cite{kerbl_3d_2023} on the Block\_A images of MatrixCity and evaluate the released models of CityGaussianV2, which reports a higher mesh F1 on GauU-Scene than 2DGS~\cite{huang_2dgs_2024} and CityGaussian~\cite{liu_citygs_2024}.
Of those that simplify an existing 3DGS model, NanoGS~\cite{xiong_nanogs_2026}, LightGaussian~\cite{fan_lightgaussian_nodate} and PUP 3D-GS~\cite{hanson_pup_2025} each bring the same converted 3DGS model down to our number of Gaussians, and the last two fine-tune on captured images after pruning.
All simplified models thus start from the same measured geometry, so the geometry they lose is lost in the simplification.
LightGaussian$^{\dagger}$ and PUP 3D-GS$^{\dagger}$ also follow the route of LiDAR-based 3DGS pipelines~\cite{zhou_drivinggaussian_2024,lim_lidar3dgs_2024,weng_structuredligs_2026}, which optimize point-cloud-initialized Gaussians against captured images, here from the full converted model.
Since their released code cannot hold a converted model of this size in memory, these two score the Gaussians in spatial tiles (marked $^{\dagger}$); on a sub-block, the tiled scoring keeps nearly the same Gaussians as the original code (see the supplementary material).

\paragraph{Implementation details.}\label{sec:impl} The conversion computes each covariance from $k=30$ neighbors with $\gamma=0.6$. Merging uses $\kappa=0.6$ and keeps $\rho=75\%$ of the Gaussians per round until 1\% of the converted 3DGS model remains (17 rounds on MatrixCity Aerial).
Without captured images, distillation runs for 8k steps from teacher renderings at the camera poses of the training images; with captured images, it runs for 20k steps from them, with $\bar{s}=10$\,m in the scale regularizer.
All experiments use one Quadro RTX 6000 GPU, and merging runs on the CPU.

\subsection{Comparison with Existing Methods}
\label{sec:exp_comparison}

\paragraph{Geometry.} Since the point cloud already measures the scene, the first question is how much of this geometry each model keeps at the same number of Gaussians (\cref{tab:main}).
Without any training, the converted 3DGS model reaches an AbsRel about 17 times lower than 3DGS trained on the aerial images, and lower than CityGaussianV2 on both datasets.
PUP 3D-GS$^{\dagger}$~\cite{hanson_pup_2025} starts from the same converted model and refines it against the captured images without depth supervision.
This refinement re-estimates the geometry, and the AbsRel of PUP 3D-GS$^{\dagger}$ rises to more than twice that of the converted model on both datasets (\cref{fig:qual}).
After merging to 1\% and distillation, both simplified models keep the geometry of the converted model on MatrixCity Aerial, and the one with captured images more than halves the depth error of the converted model.
By contrast, LightGaussian$^{\dagger}$~\cite{fan_lightgaussian_nodate} removes nearly all roofs.
Merging without training as well, NanoGS likewise keeps the depth error of the converted model, yet fewer of its pixels lie within 5\% of the true depth.
On GauU-Scene, the model without captured images matches or exceeds every baseline in AbsRel and mesh F1.
In mesh F1, both simplified models exceed every baseline on MatrixCity Aerial.

\begin{figure*}[t]
\centering
\includegraphics[width=\textwidth]{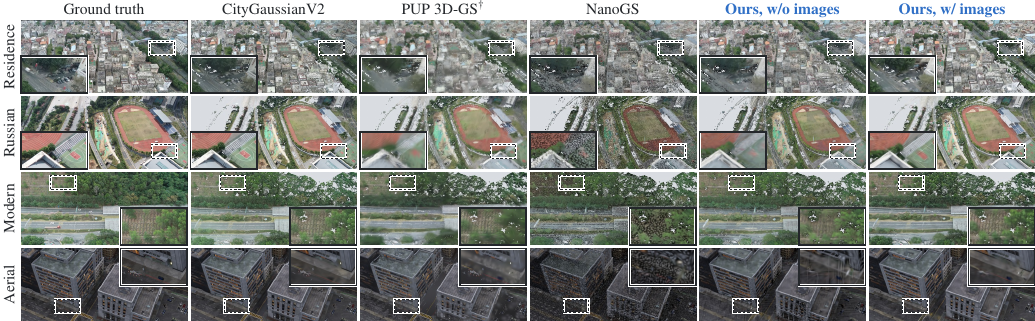}
\caption{Renderings at the views and boxes of \cref{fig:qual}; light gray pixels have no ground truth. Distillation restores the texture of GARD-GS, whereas NanoGS keeps the averaged colors; CityGaussianV2 and PUP 3D-GS$^{\dagger}$ render plausible images even with wrong depth.}
\label{fig:qual_render}
\end{figure*}

\paragraph{Appearance.} Merging averages the colors, so the next question is how much appearance distillation restores.
NanoGS keeps the averaged colors and renders below the converted 3DGS model on MatrixCity Aerial, whereas distillation from teacher renderings alone lifts PSNR above it, and captured images add a further 3.9\,dB while also lowering the depth error.
PUP 3D-GS$^{\dagger}$ renders on par but has more than five times this depth error, and CityGaussianV2 reaches a higher PSNR with 5.8 times as many Gaussians, at 4.5 times the depth error.
On GauU-Scene, captured images raise PSNR by 4.0\,dB on average to the highest PSNR of all methods, with an AbsRel about 1.8 times lower than that of CityGaussianV2 (\cref{fig:qual_render}).

\begin{table}[t]
\caption{Ablation study on MatrixCity Aerial (0.66M Gaussians, w/o captured images). Each variant below the line changes one design decision; all end with distillation.}
\label{tab:ablation}
\centering
\footnotesize
\sbox0{%
\begin{tabular}{@{}lccc@{}}
\toprule
Variant & AbsRel$\downarrow$ & $\delta_{1.05}$$\uparrow$ & PSNR$\uparrow$ \\
\midrule
Converted model (66.2M) & 0.0244 & 0.937 & 19.56 \\
GARD-GS (17 rounds, 8k steps) & \textbf{0.0198} & \textbf{0.944} & \textbf{21.36} \\
\midrule
One merging round (8k steps) & 0.0260 & 0.913 & 20.58 \\
Fitting after every round (34k steps) & 0.0204 & 0.941 & 20.64 \\
\bottomrule
\end{tabular}}
\typeout{TABWIDTH tab:ablation \the\wd0\space\the\linewidth}
\ifdim\wd0>\linewidth\resizebox{\linewidth}{!}{\usebox0}\else\usebox0\fi
\end{table}

\begin{figure}[t]
\centering
\includegraphics[width=\linewidth]{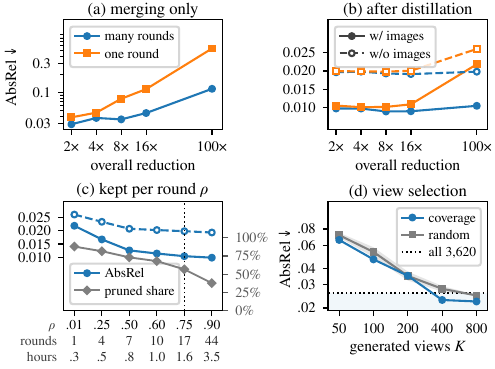}
\caption{Many rounds against one on MatrixCity Aerial, ending at 661,757 Gaussians. (a) Merging only: against the overall reduction, many rounds give a lower AbsRel from 2$\times$ to 100$\times$. (b) The same after distillation. (c) AbsRel and the pruned share of the reduction against $\rho$; dotted: the $\rho$ we use. (d) AbsRel against the number $K$ of generated views, chosen by coverage or at random (band: $\pm1$ std over three seeds); dotted: all 3,620 candidates.}
\label{fig:multiround}
\end{figure}

\subsection{Ablation Studies}
\label{sec:exp_ablation}

Each ablation changes one design decision of \cref{sec:method} on MatrixCity Aerial, whose synthetic ground truth is exact at every pixel, simplifying the converted model of $6.6\times10^7$ Gaussians to the same 661,757 Gaussians as in \cref{tab:main}.
The ablations with captured images follow this design in the supplementary material.

\paragraph{Fitting between rounds.} Pruning methods for trained 3DGS models fine-tune between rounds to repair what each round removes.
A converted model, however, already takes its geometry from the point cloud, so such fitting has little to repair.
Fitting for 2k steps after each of rounds 4--16 and 8k after the last (34k steps in total) lowers the AbsRel no further than a single fitting after the last round (\cref{tab:ablation}); the same holds with captured images (0.0100 against 0.0098 for one 20k-step distillation; \cref{fig:pain}c).
Deferring all optimization to one final distillation thus reaches the same geometric accuracy with far fewer steps.

\paragraph{Multi-round merging.}\label{sec:exp_multiround} The number of rounds decides how much of the simplification falls to pruning.
A single round to 1\% leaves most of it to pruning, which opens holes across whole facades (\cref{fig:pain}b); even after distillation, its AbsRel stays about a third higher than with 17 rounds without captured images (\cref{tab:ablation}) and twice as high with them at 8k steps (0.0218 against 0.0105).
As the fraction $\rho$ kept per round rises, the depth error falls until 75\% and barely changes beyond it, while the merging time keeps growing and the share of the reduction left to pruning falls from 88\% for one round to 57\% at $\rho=75\%$ (\cref{fig:multiround}c).
At a fixed final count, many rounds give the lower AbsRel before distillation from 2$\times$ on, by a margin that widens with the overall ratio (\cref{fig:multiround}a), and after distillation the margin is largest at 100$\times$ (\cref{fig:multiround}b), so many rounds matter most for our operator at the high ratios that dense point clouds require.

\paragraph{Scale regularizer.} Captured images can show content beyond the point cloud.
With captured images and 8k steps, the scale regularizer lowers the AbsRel on the aerial views of MatrixCity from 0.0116 to 0.0105.

\paragraph{Views without camera poses.} Captured images supervise distillation wherever they exist, and their camera poses place the teacher renderings when the images themselves are not used.
With the point cloud alone, neither is available, so the teacher is rendered at views generated from the point cloud and chosen by coverage.
With 400 such views, the distilled model reaches an AbsRel of 0.0230 at 8k steps, lower than with all 3,620 candidates and about half that of CityGaussianV2 trained on the captured images; at this budget, selecting by coverage gives an AbsRel nearly a fifth lower than random selection (\cref{fig:multiround}d).

\section{Conclusion}
\label{sec:conclusion}

GARD-GS converts a dense point cloud into a 3DGS model without training, simplifies this model first by training-free merging over many rounds, and optimizes it only once, in a distillation that restores its appearance.
At 1\% of the Gaussians, the simplified models keep the measured geometry, with a lower depth error than 3DGS models trained on images, including CityGaussianV2 on both benchmarks.
Without captured images, all supervision comes from the point cloud: surfaces the scanner misses, such as water and glass, stay empty, and the appearance is bounded by the point colors.
Both benchmarks are aerial; ground-level captures, whose images extend beyond the point cloud to sky and distant background, remain future work.

\FloatBarrier

\section*{LLM Usage Statement}

The authors take full responsibility for all submitted content, regardless of any LLM or AI-tool use. The tools used were ChatGPT (OpenAI) and Claude (Anthropic), and the full manuscript was given to both to obtain reviewer-style feedback. The highest level of LLM or AI-tool involvement in each part of the work was as follows. \emph{Assist}, editing or refining author-produced content: writing and editing the text (language, restructuring, drafting prose); related work and citations; experiments (design, execution, analysis of results). \emph{Generate}, the authors reviewed, edited or approved tool-generated work, through one-shot or interactive use: data (synthetic data, dataset creation, annotation); research ideas, hypotheses, claims or conclusions. \emph{Agent}, the tool carried out multi-step tasks autonomously, with limited step-by-step human direction: code and implementation; figures, diagrams and visual material.

\paragraph{Acknowledgments.} This work was conducted primarily during the first author's internship at Jishu Technology Co., Ltd. We thank the company for its research and computational support.
{
    \small
    \FloatBarrier
    \makeatletter\newcounter{gardbib}\let\gard@bibitem\bibitem
    \renewcommand{\bibitem}{\stepcounter{gardbib}\ifnum\value{gardbib}=66\relax\vfill\newpage\fi\gard@bibitem}\makeatother
    \bibliographystyle{ieeenat_fullname}
    \bibliography{main}
}

\clearpage
\providecommand{\suppcite}[2]{\hyperlink{suppref:#1}{[#2]}}
\nocite{ye_gsplat_2025}%
\maketitlesupplementary

\cref{sec:suppl_protocol} completes the evaluation protocol of \cref{sec:exp_setup}, and \cref{sec:suppl_perscene} gives the full results of \cref{tab:main}.
\cref{sec:suppl_storage} follows multi-round merging over a range of storage sizes, and \cref{sec:suppl_design} follows the single distillation with captured images and at views generated without camera poses.
\cref{sec:suppl_scan} compares converting the scan with using it inside photometric training, as an initialization or as a depth loss.
\cref{sec:suppl_qual} shows further test views, \cref{sec:suppl_discussion} discusses an extension of the distillation, and \cref{sec:suppl_impl} collects the implementation details.

\section{Evaluation Protocol}
\label{sec:suppl_protocol}

\paragraph{Datasets.} MatrixCity Aerial~\cite{li_matrixcity_2023} provides 930 training and 163 test views of Block\_A, and of its point cloud of $8.8\times10^7$ points, the $6.6\times10^7$ that remain after removing isolated points each become one Gaussian of the converted 3DGS model.
The airborne LiDAR point clouds of GauU-Scene~\cite{xiong_gauuscene_2024} hold tens of millions of points per scene, with 643 to 995 aligned drone images for training and 72 to 111 held out for testing.

\paragraph{Depth metrics and seeds.} The depth metrics in the main paper use the expected depth, and so do the AbsRel and $\delta_{1.05}$~\cite{eigen_depth_2014,beche_claravid_2025} values in this supplementary material; \cref{tab:suppl_dn} and \cref{fig:suppl_storage} add the AbsRel of the median depth, from which the meshes are fused.
Merging is deterministic, and over three seeds of the distillation on MatrixCity Aerial, the AbsRel has a standard deviation of at most 0.0001 and the PSNR varies by at most 0.02\,dB, so we report single runs.
PSNR, SSIM and LPIPS are computed at full resolution without color correction; on GauU-Scene, PSNR excludes the sky, whereas SSIM and LPIPS include it.

\paragraph{Mesh extraction.} Each mesh is cropped to the region that the training views observe and aligned to the ground truth with ICP, except the mesh of 3DGS~\cite{kerbl_3d_2023} on MatrixCity Aerial, for which the alignment diverges; F1~\cite{knapitsch_tanks_2017} uses the distance threshold of CityGaussianV2~\cite{liu_citygsv2_2025}, 1.5 times the mean nearest-neighbor spacing of the ground-truth point cloud.

\paragraph{Storage.} Sizes count the stored parameters of the Gaussians, without the empty normals of PLY files.

\section{Full Comparison Results}
\label{sec:suppl_perscene}

On every scene of GauU-Scene, the pattern of the means in \cref{tab:main} holds (\cref{tab:suppl_gauu}): GARD-GS without captured images has a higher mesh F1 and a lower depth error than CityGaussianV2, and GARD-GS with captured images has a lower depth error and the highest PSNR of all methods, both at a twentieth of the storage of CityGaussianV2 or less.
On MatrixCity Aerial (\cref{tab:suppl_full}), both simplified versions of GARD-GS exceed every baseline in precision as well as in recall, so their meshes are at once more accurate and more complete.

\begin{table*}[t]
\caption{Per-scene results on GauU-Scene. F1, AbsRel, $\delta_{1.05}$ and PSNR average to the GauU-Scene columns of \cref{tab:main}. $^{\ddagger}$Reported by CityGaussianV2 on A100 GPUs.}
\label{tab:suppl_gauu}
\centering
\footnotesize
\sbox0{%
\begin{tabular}{@{}clcccccccccc@{}}
\toprule
 & Method & F1$\uparrow$ & AbsRel$\downarrow$ & $\delta_{1.05}$$\uparrow$ & PSNR$\uparrow$ & SSIM$\uparrow$ & LPIPS$\downarrow$ & \#GS & Size & FPS & GPU$\cdot$h \\
\midrule
\multirow{7}{*}{\rotatebox[origin=c]{90}{Russian Building}} & CityGaussianV2 & 0.362 & 0.0449 & 0.717 & \underline{17.20} & 0.397 & \textbf{0.589} & 5.43M & 803 MB & 33.3$^{\ddagger}$ & 23.6$^{\ddagger}$ \\
\cmidrule{2-12}
 & NanoGS & 0.476 & \underline{0.0193} & 0.896 & 13.50 & 0.287 & 0.911 & 283K & 15.9 MB & 296 & 0 \\
 & LightGaussian$^{\dagger}$ & 0.060 & 0.1882 & 0.052 & 15.75 & \underline{0.445} & 0.918 & 283K & 19.3 MB & 104 & 0.29 \\
 & PUP 3D-GS$^{\dagger}$ & 0.200 & 0.0458 & 0.704 & 17.10 & 0.438 & 0.818 & 283K & 19.3 MB & 115 & 1.16 \\
\cmidrule{2-12}
 & Ours, converted & \textbf{0.567} & \textbf{0.0169} & \textbf{0.912} & 15.03 & 0.242 & \underline{0.728} & 28.3M & 1.59 GB & 28.6 & 0.01 \\
 & Ours, w/o captured images & \underline{0.480} & \underline{0.0193} & \underline{0.903} & 15.75 & 0.343 & 0.862 & 283K & 15.9 MB & 162 & 0.20 \\
 & Ours, w/ captured images & 0.353 & 0.0266 & 0.866 & \textbf{20.53} & \textbf{0.506} & 0.739 & 283K & 15.9 MB & 91 & 0.36 \\
\midrule
\multirow{7}{*}{\rotatebox[origin=c]{90}{Residence}} & CityGaussianV2 & 0.403 & 0.0818 & 0.576 & 15.41 & 0.319 & \textbf{0.695} & 8.03M & 1189 MB & 45.5$^{\ddagger}$ & 24.1$^{\ddagger}$ \\
\cmidrule{2-12}
 & NanoGS & \underline{0.435} & \underline{0.0331} & 0.810 & 13.98 & 0.286 & 0.925 & 985K & 55.2 MB & 229 & 0 \\
 & LightGaussian$^{\dagger}$ & 0.053 & 0.3975 & 0.033 & 13.47 & 0.360 & 0.936 & 985K & 67.0 MB & 126 & 0.45 \\
 & PUP 3D-GS$^{\dagger}$ & 0.243 & 0.0907 & 0.432 & \underline{15.79} & \underline{0.364} & 0.885 & 985K & 67.0 MB & 140 & 1.38 \\
\cmidrule{2-12}
 & Ours, converted & \textbf{0.529} & \textbf{0.0293} & \textbf{0.833} & 14.76 & 0.242 & \underline{0.756} & 98.5M & 5.52 GB & 20.8 & 0.02 \\
 & Ours, w/o captured images & 0.430 & \underline{0.0331} & \underline{0.814} & 15.30 & 0.322 & 0.906 & 985K & 55.3 MB & 149 & 0.31 \\
 & Ours, w/ captured images & 0.384 & 0.0420 & 0.767 & \textbf{18.65} & \textbf{0.413} & 0.828 & 985K & 55.3 MB & 99 & 0.40 \\
\midrule
\multirow{7}{*}{\rotatebox[origin=c]{90}{Modern Building}} & CityGaussianV2 & 0.408 & 0.0644 & 0.630 & 16.41 & 0.384 & \textbf{0.637} & 7.94M & 1175 MB & 34.5$^{\ddagger}$ & 24.7$^{\ddagger}$ \\
\cmidrule{2-12}
 & NanoGS & \underline{0.476} & \underline{0.0270} & 0.866 & 15.01 & 0.346 & 0.893 & 942K & 52.8 MB & 195 & 0 \\
 & LightGaussian$^{\dagger}$ & 0.070 & 0.1597 & 0.115 & 15.87 & \underline{0.448} & 0.930 & 942K & 64.1 MB & 98 & 0.40 \\
 & PUP 3D-GS$^{\dagger}$ & 0.222 & 0.0578 & 0.600 & \underline{17.42} & 0.439 & 0.825 & 942K & 64.1 MB & 111 & 1.21 \\
\cmidrule{2-12}
 & Ours, converted & \textbf{0.523} & \textbf{0.0244} & \textbf{0.879} & 16.01 & 0.309 & \underline{0.689} & 94.2M & 5.28 GB & 13.7 & 0.02 \\
 & Ours, w/o captured images & \underline{0.476} & 0.0271 & \underline{0.868} & 16.67 & 0.407 & 0.872 & 942K & 52.8 MB & 123 & 0.26 \\
 & Ours, w/ captured images & 0.413 & 0.0354 & 0.825 & \textbf{20.65} & \textbf{0.503} & 0.763 & 942K & 52.8 MB & 85 & 0.40 \\
\bottomrule
\end{tabular}}
\typeout{TABWIDTH tab:suppl_gauu \the\wd0\space\the\linewidth}
\ifdim\wd0>\linewidth\resizebox{\linewidth}{!}{\usebox0}\else\usebox0\fi
\end{table*}

\begin{table*}[t]
\caption{Full metrics on MatrixCity Aerial; precision and recall belong to the mesh F1.}
\label{tab:suppl_full}
\centering
\footnotesize
\sbox0{%
\begin{tabular}{@{}lcccccccc@{}}
\toprule
Method & F1$\uparrow$ & Prec.$\uparrow$ & Recall$\uparrow$ & AbsRel$\downarrow$ & $\delta_{1.05}$$\uparrow$ & PSNR$\uparrow$ & SSIM$\uparrow$ & LPIPS$\downarrow$ \\
\midrule
3DGS & 0.291 & 0.261 & 0.329 & 0.4168 & 0.281 & 20.75 & 0.671 & 0.576 \\
CityGaussianV2 & 0.527 & 0.435 & 0.667 & 0.0438 & 0.706 & \textbf{27.43} & \textbf{0.821} & \textbf{0.284} \\
\midrule
NanoGS & 0.617 & 0.497 & 0.812 & 0.0208 & 0.897 & 17.34 & 0.354 & 0.707 \\
LightGaussian$^{\dagger}$ & 0.029 & 0.028 & 0.029 & 0.2265 & 0.173 & 18.45 & 0.575 & 0.855 \\
PUP 3D-GS$^{\dagger}$ & 0.466 & 0.360 & 0.664 & 0.0539 & 0.646 & 25.02 & \underline{0.747} & \underline{0.410} \\
\midrule
Ours, converted & \underline{0.775} & 0.683 & \textbf{0.896} & 0.0244 & 0.937 & 19.56 & 0.323 & 0.561 \\
Ours, w/o captured images & \textbf{0.777} & \textbf{0.710} & \underline{0.858} & \underline{0.0198} & \underline{0.944} & 21.36 & 0.566 & 0.578 \\
Ours, w/ captured images & 0.757 & \underline{0.691} & 0.836 & \textbf{0.0098} & \textbf{0.968} & \underline{25.28} & 0.724 & 0.470 \\
\bottomrule
\end{tabular}}
\typeout{TABWIDTH tab:suppl_full \the\wd0\space\the\textwidth}
\ifdim\wd0>\textwidth\resizebox{\textwidth}{!}{\usebox0}\else\usebox0\fi
\end{table*}

\section{Storage Sweep}
\label{sec:suppl_storage}

\cref{tab:main} shows that multi-round merging keeps the measured geometry at 1\% of the converted 3DGS model, and this section follows it over the storage size, against the lightweight methods at the same storage and against larger models that learn their geometry from images or with the scan depth.

\begin{figure*}[tp]
\centering
\includegraphics[width=\textwidth]{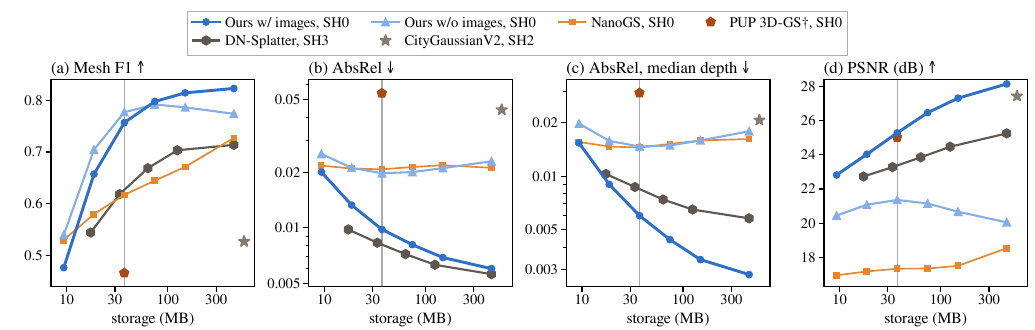}
\caption{Geometry and appearance against storage on MatrixCity Aerial. The gray vertical line marks the storage of GARD-GS in \cref{tab:main}, and the rightmost points of GARD-GS and NanoGS ($12.1N$) match the storage of the native arm of DN-Splatter. $N=661{,}757$.}
\label{fig:suppl_storage}
\end{figure*}

Each size changes only the target number of Gaussians in the recipe of \cref{tab:main}, and NanoGS~\cite{xiong_nanogs_2026} merges the converted 3DGS model to the same numbers.
Among the lightweight methods, GARD-GS forms the Pareto front in all four metrics (\cref{fig:suppl_storage}), and with captured images it improves in every metric with each increase of the storage.
Among all models drawn, GARD-GS with captured images also forms the PSNR front at every storage, and at its largest size it renders better than CityGaussianV2, which stores more.
Already at $N$, GARD-GS has a higher mesh F1 than both the native arm of DN-Splatter~\suppcite{dnsplatter}{75} and CityGaussianV2, at a twelfth and a fifteenth of their storage.
The storage of GARD-GS can thus be chosen along this front.

\section{Distillation Ablations}
\label{sec:suppl_design}

A single distillation restores the appearance of the simplified model, from captured images or, without them, from renderings of the converted 3DGS model (\cref{sec:distillation}).
This section carries the ablations of \cref{sec:exp_ablation} over to captured images and details how GARD-GS chooses the views to render when no camera poses exist.

\paragraph{Ablations with captured images.} Multi-round merging remains the most important component, as reducing in one round roughly doubles the depth error (\cref{tab:suppl_ablation_delta}); the scale regularizer lowers the depth error at no cost in rendering.

\begin{table}[t]
\caption{Ablation study with captured images, on MatrixCity Aerial (661,757 Gaussians) as in \cref{tab:ablation}.}
\label{tab:suppl_ablation_delta}
\centering
\footnotesize
\sbox0{%
\begin{tabular}{@{}lccc@{}}
\toprule
Variant & AbsRel$\downarrow$ & $\delta_{1.05}$$\uparrow$ & PSNR$\uparrow$ \\
\midrule
GARD-GS, 8k & 0.0105 & 0.962 & 24.53 \\
GARD-GS, 20k (\cref{tab:main}) & \textbf{0.0098} & \textbf{0.968} & \textbf{25.28} \\
one round, 8k & 0.0218 & 0.911 & 22.53 \\
w/o scale regularizer, 8k & 0.0116 & 0.957 & 24.53 \\
\bottomrule
\end{tabular}}
\typeout{TABWIDTH tab:suppl_ablation_delta \the\wd0\space\the\linewidth}
\ifdim\wd0>\linewidth\resizebox{\linewidth}{!}{\usebox0}\else\usebox0\fi
\end{table}

\begin{figure}[tp]
\centering
\IfFileExists{fig/fig_suppl_views.pdf}{\includegraphics[width=\linewidth]{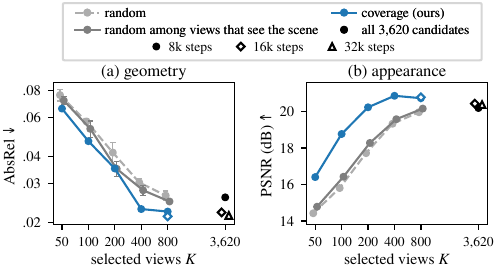}}{\fbox{\parbox[c][0.45\linewidth][c]{0.95\linewidth}{\centering\missing{} \texttt{fig/fig\_suppl\_views.pdf}}}}
\caption{View selection without captured images on MatrixCity Aerial: $K$ of 3,620 generated candidates distill the same merged model (661,757 Gaussians). Random selection: mean and standard deviation over three seeds; coverage-based selection is deterministic.}
\label{fig:suppl_views}
\end{figure}

\paragraph{Views without camera poses.} Without camera poses, the teacher is rendered at views generated from the point cloud (\cref{sec:distillation}), and GARD-GS chooses $K$ of them by how much of the surface each one adds.
What a view shows is described by observation bins: a bin $b$ groups the pixels that show the same surface voxel from a similar direction and distance, and $\omega_{vb}$ is the number of pixels of candidate $v$ in $b$.
A set $\mathcal{U}$ of views covers bin $b$ with its pixels there, counted in units of a typical view of the bin, the mean $m_b$ of $\omega_{vb}$ over the candidates that see $b$:
\begin{equation}
c_b(\mathcal{U})=\frac{1}{m_b}\sum_{v\in\mathcal{U}}\omega_{vb}.
\end{equation}
The first view of a bin adds more than a further one, so coverage counts with diminishing returns as $1-e^{-c_b/\beta}$, with $\beta=0.5$.
Weighting each bin by $W_b=\sum_v\omega_{vb}$, the number of its pixels over all candidates, gives the coverage of the set,
\begin{equation}
F(\mathcal{U})=\sum_b W_b\big(1-e^{-c_b(\mathcal{U})/\beta}\big),
\end{equation}
and GARD-GS builds $\mathcal{U}$ greedily, adding at each step the candidate that raises $F$ most, so that every new view goes mostly to bins that the selected views have not covered yet.
Random selection, here and in \cref{fig:multiround}(d), draws among the candidates that see the scene.
Coverage-based selection gives the better rendering at every budget (\cref{fig:suppl_views}), and the lower depth error at small budgets and from $K=400$ on.
At the same distillation steps, the $K=400$ views that GARD-GS selects give a lower depth error than all candidates; with more steps, 800 views selected by coverage reach the depth error of all candidates in half the steps and render better.
Selecting views by coverage thus makes distillation more efficient for a given budget of views and steps.

\section{Comparison with Scan-Based Training}
\label{sec:suppl_scan}

When a scan is available, 3DGS pipelines use it to help produce a model rather than to become one (\cref{sec:intro}): the scan initializes the Gaussians, or its depth enters the photometric loss, and in both cases the optimization still estimates the geometry.
GARD-GS instead converts the scan into the model, so the geometry is measured from the start and distillation refines it.
Keeping the measured geometry is the aim of the task, so the scan is also the reference for the geometry, and every method in this section starts from this same scan; the comparison thus isolates how each method uses it.

\subsection{Scan Initialization versus Conversion}
\label{sec:suppl_sameinput}

We isolate the start of training: from the same point cloud, at the same number of Gaussians and with the same captured images, losses and 20k steps, a standard 3DGS initialized from the voxel-downsampled point cloud is compared with GARD-GS, which starts from the converted and merged model (\cref{sec:suppl_impl}).
The comparison is repeated with the photometric loss alone and with a scan-depth loss added, the two ways in which 3DGS pipelines use a scan (\cref{tab:suppl_same_aerial}).
With the photometric loss alone, GARD-GS halves the depth error of the initialized model on MatrixCity Aerial, lowers it on every scene of GauU-Scene and has the higher mesh F1 on all four scenes.
Initializing from the scan is thus not enough to keep its geometry: under a photometric loss that leaves the depth weakly constrained (\cref{sec:method}), the isotropic, semi-transparent Gaussians of the initialized model learn their shapes and opacities from the images, free to move their surfaces away from the points that measured them. GARD-GS instead starts distillation from Gaussians that already lie along the measured surfaces, and its scale regularizer keeps them there (\cref{sec:distillation}), so distillation refines the appearance rather than re-estimating the geometry.
With the scan depth added, the converted start gives the higher mesh F1 and PSNR on all four scenes, so conversion keeps its advantage when the scan also supervises the depth.

\begin{table*}[t]
\caption{Scan initialization against conversion from the same point cloud on MatrixCity Aerial and the three scenes of GauU-Scene, at the same number of Gaussians and 20k steps; the two rows of each pair share their losses. The learning rate and loss weight of the initialized models and of conversion with the scan-depth loss are selected on 133 held-out training views of MatrixCity Aerial.}
\label{tab:suppl_same_aerial}
\centering
\footnotesize
\sbox0{%
\begin{tabular}{@{}llccccccccccccc@{}}
\toprule
 & & \multicolumn{3}{c}{MatrixCity Aerial} & \multicolumn{3}{c}{Russian Building} & \multicolumn{3}{c}{Residence} & \multicolumn{3}{c}{Modern Building} \\
\cmidrule(lr){3-5}\cmidrule(lr){6-8}\cmidrule(lr){9-11}\cmidrule(lr){12-14}
Loss & Start & AbsRel$\downarrow$ & PSNR$\uparrow$ & F1$\uparrow$ & AbsRel$\downarrow$ & PSNR$\uparrow$ & F1$\uparrow$ & AbsRel$\downarrow$ & PSNR$\uparrow$ & F1$\uparrow$ & AbsRel$\downarrow$ & PSNR$\uparrow$ & F1$\uparrow$ \\
\midrule
\multirow{2}{*}{Photometric} & Scan initialization & 0.0203 & 24.53 & 0.463 & 0.0455 & \textbf{21.06} & 0.271 & 0.0533 & \textbf{18.80} & 0.339 & 0.0654 & \textbf{21.17} & 0.378 \\
 & Conversion (GARD-GS) & \textbf{0.0098} & \textbf{25.28} & \textbf{0.757} & \textbf{0.0266} & 20.53 & \textbf{0.353} & \textbf{0.0420} & 18.65 & \textbf{0.384} & \textbf{0.0354} & 20.65 & \textbf{0.413} \\
\midrule
\multirow{2}{*}{\shortstack[l]{Photometric\\+ scan depth}} & Scan initialization & \textbf{0.0071} & 24.47 & 0.752 & 0.0184 & 20.50 & 0.437 & 0.0343 & 18.46 & 0.414 & \textbf{0.0270} & 20.75 & 0.428 \\
 & Conversion & \textbf{0.0071} & \textbf{25.21} & \textbf{0.774} & \textbf{0.0183} & \textbf{20.61} & \textbf{0.458} & \textbf{0.0340} & \textbf{18.64} & \textbf{0.425} & 0.0271 & \textbf{20.83} & \textbf{0.437} \\
\bottomrule
\end{tabular}}
\typeout{TABWIDTH tab:suppl_same_aerial \the\wd0\space\the\textwidth}
\ifdim\wd0>\textwidth\resizebox{\textwidth}{!}{\usebox0}\else\usebox0\fi
\end{table*}

\subsection{Published LiDAR-Supervised Methods}
\label{sec:suppl_published}

DN-Splatter~\suppcite{dnsplatter}{75} and LetsGo~\suppcite{letsgo}{74} add LiDAR depth to the photometric loss of 3DGS, as the initialized model with the scan-depth loss does, and \cref{sec:suppl_impl} describes their adaptation.
We compare them with GARD-GS on MatrixCity Aerial at the same storage and per block.

\paragraph{DN-Splatter.} At every matched size (\cref{tab:suppl_dn}), GARD-GS with captured images has the lower error of the median depth, and with it the higher mesh F1 and the higher PSNR.
At $2N$, a sixth of the storage of the native arm, GARD-GS already exceeds it in mesh F1 and PSNR, and at equal storage it leads in the median depth, the mesh F1 and the PSNR.
At the same storage, GARD-GS thus gives the more accurate mesh and the better rendering without any depth loss.

\begin{table*}[t]
\caption{DN-Splatter against GARD-GS with captured images on MatrixCity Aerial, $N=661{,}757$. The matched arms cap DN-Splatter at the storage of GARD-GS at $N/2$, $N$ and $2N$, and GARD-GS at $12.1N$ matches the storage of the native arm. SH: degree of the spherical harmonics; AbsRel$_{\mathrm{med}}$: AbsRel of the median depth.}
\label{tab:suppl_dn}
\centering
\footnotesize
\sbox0{%
\begin{tabular}{@{}lcccccccc@{}}
\toprule
Method & \#GS & Size & SH & AbsRel$\downarrow$ & AbsRel$_{\mathrm{med}}$$\downarrow$ & $\delta_{1.05}$$\uparrow$ & PSNR$\uparrow$ & F1$\uparrow$ \\
\midrule
DN-Splatter, matched $N/2$ & 73,116 & 17.26 MB & 3 & \textbf{0.0098} & 0.0103 & \textbf{0.977} & 22.73 & 0.544 \\
Ours, w/ captured images, $N/2$ & 330,878 & 18.62 MB & 0 & 0.0133 & \textbf{0.0090} & 0.952 & \textbf{24.03} & \textbf{0.657} \\
\midrule
DN-Splatter, matched $N$ & 141,458 & 33.38 MB & 3 & \textbf{0.0083} & 0.0087 & \textbf{0.981} & 23.28 & 0.618 \\
Ours, w/ captured images, $N$ & 661,757 & 37.15 MB & 0 & 0.0098 & \textbf{0.0060} & 0.968 & \textbf{25.28} & \textbf{0.757} \\
\midrule
DN-Splatter, matched $2N$ & 268,860 & 63.45 MB & 3 & \textbf{0.0072} & 0.0074 & \textbf{0.985} & 23.85 & 0.669 \\
Ours, w/ captured images, $2N$ & 1,323,514 & 74.21 MB & 0 & 0.0081 & \textbf{0.0044} & 0.974 & \textbf{26.46} & \textbf{0.797} \\
\midrule
DN-Splatter, native & 1,900,903 & 448.61 MB & 3 & \textbf{0.0056} & 0.0058 & \textbf{0.990} & 25.25 & 0.714 \\
Ours, w/ captured images, $12.1N$ & 8,010,893 & 448.70 MB & 0 & 0.0060 & \textbf{0.0028} & 0.980 & \textbf{28.14} & \textbf{0.823} \\
\bottomrule
\end{tabular}}
\typeout{TABWIDTH tab:suppl_dn \the\wd0\space\the\textwidth}
\ifdim\wd0>\textwidth\resizebox{\textwidth}{!}{\usebox0}\else\usebox0\fi
\end{table*}

\paragraph{LetsGo$^{\dagger}$.} LetsGo$^{\dagger}$ trains the scene in blocks, and a single block already holds two to four times as many Gaussians as GARD-GS uses for the whole scene and stores 8 to 16 times as much.
Yet one GARD-GS model of the whole scene renders every block with a higher PSNR than the LetsGo$^{\dagger}$ model trained for that block alone, and on blocks 10 and 42 it also has the lower depth error, about half that of LetsGo$^{\dagger}$ on block 10 (\cref{tab:suppl_lg}).
Converting the scan thus gives one model of the whole scene that renders better than per-block training with the scan depth, with a fraction of the Gaussians and the storage of a single block.

\begin{table}[tbp]
\caption{LetsGo$^{\dagger}$ on three blocks of MatrixCity Aerial against one GARD-GS model of the whole scene ($N=661{,}757$, with captured images), evaluated inside each block on the test views that see it. LetsGo$^{\dagger}$ stores spherical harmonics of degree 3, GARD-GS of degree 0.}
\label{tab:suppl_lg}
\centering
\footnotesize
\sbox0{%
\begin{tabular}{@{}llrrrr@{}}
\toprule
Block (views) & Method & AbsRel$\downarrow$ & $\delta_{1.05}$$\uparrow$ & PSNR$\uparrow$ & \#GS \\
\midrule
\multirow{2}{*}{10 (35)} & LetsGo$^{\dagger}$ & 0.0107 & 0.927 & 26.29 & 2,553,087 \\
 & GARD-GS & \textbf{0.0057} & \textbf{0.990} & \textbf{27.19} & 661,757 \\
\midrule
\multirow{2}{*}{22 (35)} & LetsGo$^{\dagger}$ & \textbf{0.0117} & \textbf{0.936} & 22.09 & 1,225,210 \\
 & GARD-GS & 0.0139 & 0.928 & \textbf{24.85} & 661,757 \\
\midrule
\multirow{2}{*}{42 (21)} & LetsGo$^{\dagger}$ & 0.0335 & 0.737 & 21.77 & 2,083,807 \\
 & GARD-GS & \textbf{0.0222} & \textbf{0.878} & \textbf{24.08} & 661,757 \\
\bottomrule
\end{tabular}}
\typeout{TABWIDTH tab:suppl_lg \the\wd0\space\the\linewidth}
\ifdim\wd0>\linewidth\resizebox{\linewidth}{!}{\usebox0}\else\usebox0\fi
\end{table}

\section{Additional Qualitative Results}
\label{sec:suppl_qual}

\begin{figure*}[tp]
\centering
\includegraphics[width=\textwidth]{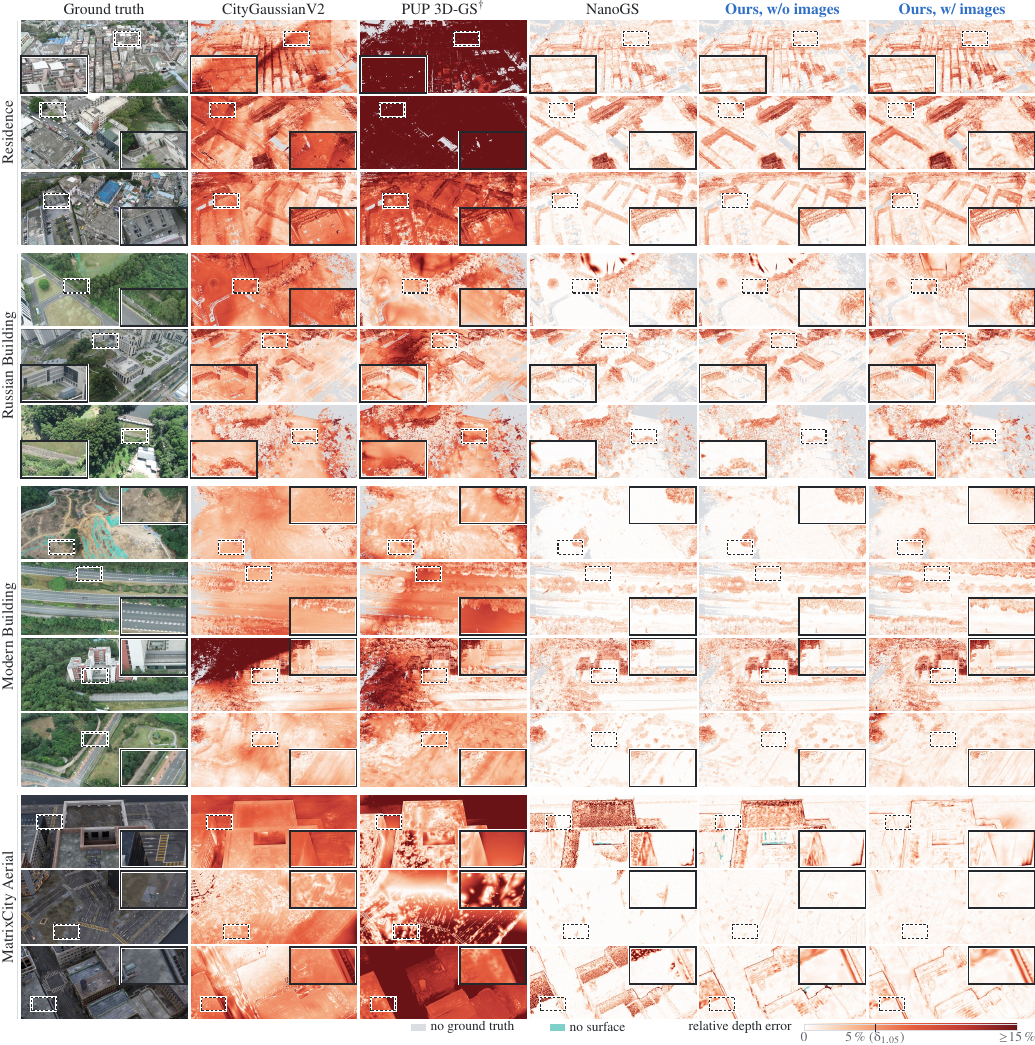}
\caption{Per-pixel relative error of the expected depth at further test views of GauU-Scene and MatrixCity Aerial, capped at 15\%; dashed boxes are enlarged 2.7 times.}
\label{fig:suppl_qual_depth}
\end{figure*}

\begin{figure*}[tp]
\centering
\includegraphics[width=\textwidth]{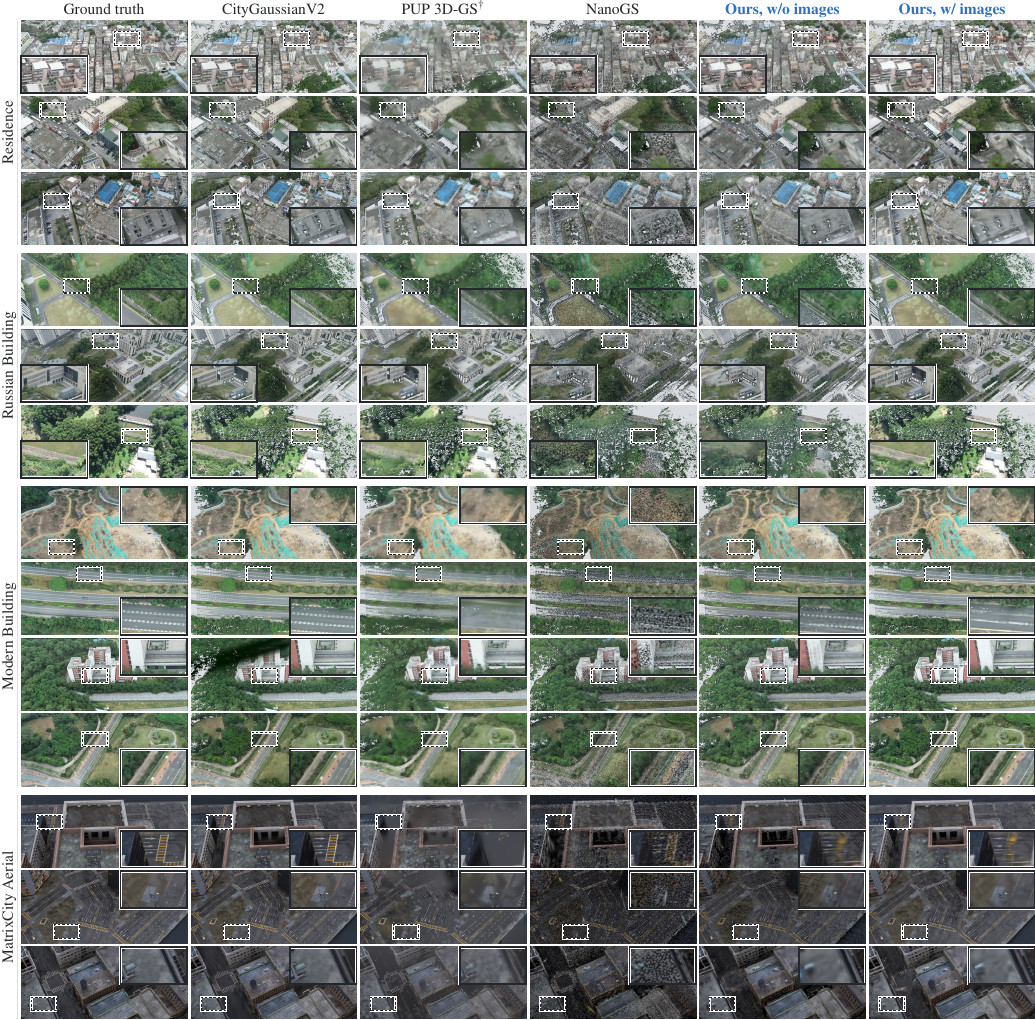}
\caption{Renderings at the views and boxes of \cref{fig:suppl_qual_depth}.}
\label{fig:suppl_qual_render}
\end{figure*}

\cref{fig:suppl_qual_depth,fig:suppl_qual_render} show further test views of every scene.
CityGaussianV2 and PUP 3D-GS$^{\dagger}$~\cite{hanson_pup_2025} leave the scanned surface over large areas, including roads and grassland, whereas NanoGS and both versions of GARD-GS stay close to it; CityGaussianV2 hides its depth error behind plausible renderings, and PUP 3D-GS$^{\dagger}$ renders blurred in several views.
NanoGS renders coarse, speckled colors, which distillation smooths, and with captured images GARD-GS restores texture of the ground truth such as lane markings.

\section{Discussion}
\label{sec:suppl_discussion}

\paragraph{Supervision at generated views.} Without captured images, distillation is supervised by renderings of the converted 3DGS model (\cref{sec:distillation}), and these renderings show the scene as the scan measured it, with holes where the scan has no coverage and with the appearance of the point colors.
We see repairing these renderings with the single-step diffusion model Difix3D+~\cite{wu_difix3d_2025} before distillation as a direct way to extend the supervision beyond what the scan measured.
The geometry of the scan could constrain such a repair to leave the surface positions unchanged and correct only the appearance.

\section{Implementation Details}
\label{sec:suppl_impl}

\subsection{GARD-GS}

\paragraph{Distillation.} Unless a row names other steps, all ablations distill for 8k steps, with and without captured images.
The photometric loss is $\mathcal{L}_{\mathrm{rgb}}=0.8\,\mathcal{L}_1+0.2\,\mathcal{L}_{\text{D-SSIM}}$, and the scale regularizer has the weight $\lambda=1$.

\paragraph{Opacity of a merged Gaussian.} \cref{sec:merging} gives a merged Gaussian the mean of its members' colors with the weights $\bar{\alpha}_i$, and an opacity at which it blocks as much light as its members together.
A member $G_i$ lets a fraction $1-\alpha_i$ of the light pass over the area $A_i$ it covers in its tangent plane, taken as the product of its two largest scales.
The merged Gaussian, covering an area $A$, takes the opacity $\alpha$ for which
\begin{equation}
(1-\alpha)^{A}=\prod_{i\in\mathcal{Q}}(1-\alpha_i)^{A_i},
\end{equation}
so a merged Gaussian that also spans the gaps between its members becomes more transparent, and one that replaces overlapping members becomes more opaque.

\paragraph{Merging details.} Voxels whose colors vary strongly are left unmerged (attribute-energy threshold 0.01). The reduction in \cref{fig:multiround}(a, b) is varied at a fixed final count by voxel-downsampling the point cloud before conversion.

\subsection{Baselines}

\paragraph{CityGaussianV2 in \cref{tab:main}.} On MatrixCity Aerial, \#GS of CityGaussianV2 counts the Gaussians inside the horizontal extent of the Block\_A point cloud, and on GauU-Scene it counts the full models.

\paragraph{Simplification baselines.} NanoGS, LightGaussian~\cite{fan_lightgaussian_nodate} and PUP 3D-GS run the official code with the default parameters and fine-tuning schedules of their authors; only the number of Gaussians they keep is set to that of GARD-GS.

\paragraph{Tiled scoring of LightGaussian$^{\dagger}$ and PUP 3D-GS$^{\dagger}$.}
Whenever a model is too large for the released code to hold, LightGaussian$^{\dagger}$ and PUP 3D-GS$^{\dagger}$ score its Gaussians in spatial tiles under one global threshold.
On a sub-block of MatrixCity Aerial with 6M Gaussians, the tiled scoring agrees with the official scoring as closely as two official runs of LightGaussian agree with each other: the recall against the official run is 0.950 for LightGaussian at 90\% pruning, against 0.960 between two official runs, and 0.980 for PUP 3D-GS at 80\% pruning.

\paragraph{Baselines from the same scan.} The initialized model voxel-downsamples the point cloud to the target number of points and initializes a standard 3DGS from them (isotropic scales from the distances to the three nearest neighbors, identity rotation, opacity 0.1, the colors of the points).
The scan-depth loss is the L1 distance between the expected depth rendered at a training view and the depth of the point cloud projected to that view (nearest point per pixel, holes up to 4 pixels filled), on the valid pixels with weight 0.01 and from step 3,000 on.
Conversion with this loss starts from the merged model of GARD-GS before distillation and trains it with the recipe of GARD-GS with captured images and the same loss.

\paragraph{Published depth-supervised baselines.} DN-Splatter starts from the full input point cloud of the scene, as GARD-GS does, keeps its native densification and stores its 3DGS model with spherical harmonics of degree 3, as in the code of its authors.
We derive its normal targets from the scan depth, where the original code derives them from the rendered depth, and select the weight 0.2 of its depth loss over 0.5 on the held-out views; its capped and matched arms stop densification at the numbers of Gaussians that correspond to the stated storage.
LetsGo does not fit the full input of MatrixCity Aerial into the 24\,GB of GPU memory, so we split the scene into $5\times6=30$ spatial blocks of about $2.9\times10^6$ input points each and train three representative blocks (10, 22 and 42); the dagger marks this training in blocks.
We switch off its levels of detail, keep its spherical harmonics of degree 3, adapt its depth warping and projection to the aerial cameras, and select the weight of its depth loss between 0.5 and 2.0 on block 22.
The number of Gaussians of LetsGo$^{\dagger}$ is that of the core region of each block.

\raggedbottom
\section*{Supplementary References}
\noindent\begin{minipage}{\linewidth}
{\small
\begin{list}{}{\settowidth{\labelwidth}{[75]}\setlength{\leftmargin}{\labelwidth}\addtolength{\leftmargin}{\labelsep}\setlength{\itemsep}{0pt}\setlength{\parsep}{0pt}\setlength{\topsep}{0pt}}\interlinepenalty=10000
\item[\hypertarget{suppref:letsgo}{[74]}] Jiadi Cui, Junming Cao, Fuqiang Zhao, Zhipeng He, Yifan Chen, Yuhui Zhong, Lan Xu, Yujiao Shi, Yingliang Zhang, and Jingyi Yu. LetsGo: Large-scale garage modeling and rendering via LiDAR-assisted Gaussian primitives. \emph{ACM TOG}, 43(6):172:1--172:18, 2024.
\item[\hypertarget{suppref:dnsplatter}{[75]}] Matias Turkulainen, Xuqian Ren, Iaroslav Melekhov, Otto Seiskari, Esa Rahtu, and Juho Kannala. DN-Splatter: Depth and normal priors for Gaussian splatting and meshing. In \emph{WACV}, pages 2421--2431, 2025.
\end{list}}
\end{minipage}

\end{document}